\documentclass[sigconf,nonacm]{acmart}

\setcopyright{none}

\usepackage{makecell}
\usepackage{amsfonts}
\usepackage{multirow}
\usepackage{subcaption}
\usepackage{wrapfig}
\usepackage{dblfloatfix}
\usepackage{placeins}
\usepackage{bbm}
\usepackage{array}
\usepackage{bm}
\usepackage{colortbl}

\newcommand{\inlinesec}[1]{{\vspace{0.5em}\noindent\textbf{#1}}\hspace{0.5em}}
\newcommand{\ours}{\textsc{MedCBR}}
\newcommand{\cbp}{\textsc{CBP}}
\newcommand{\mrg}{\textsc{MRG}}

\begin{document}

\title{Aligning Medical Vision Language Models at the Capability Boundary with Entropy Adaptive Rubric Grounding}

\author{Hongyu Bai}
\affiliation{
  \institution{Shanghai Jiao Tong University}
  \city{Shanghai}
  \country{China}
}
\email{baihongy@sjtu.edu.cn}

\author{Jiaqi Zeng}
\affiliation{
  \institution{Hunan University}
  \city{Changsha}
  \country{China}
}
\email{zjqxxl@hnu.edu.cn}

\author{Chunzheng Zhu}
\authornote{Corresponding author.}
\affiliation{
  \institution{Hunan University}
  \city{Changsha}
  \country{China}
}
\email{zhuchzh@hnu.edu.cn}

\renewcommand{\shortauthors}{Bai et al.}

\begin{abstract}
Medical diagnostic reasoning requires establishing verifiable links between visual evidence and clinical knowledge, yet current medical vision language models are aligned with objectives that reward the wrong behavior. We identify two coupled failure modes. First, conventional supervised fine tuning distributes gradient uniformly across tokens, so easy majority cases dominate the update while clinically critical boundary cases, the ones where a single diagnostic decision flips the outcome, saturate prematurely and remain under trained. Second, sparse verifiable rewards supervise only whether the final answer is correct, a single bit that cannot express whether the reasoning is anatomically grounded or whether differentials follow valid clinical logic, which invites reward hacking through answer shortcuts.
We propose \ours{}, a two stage alignment framework that couples boundary aware preference optimization with rubric grounded reinforcement learning. We first construct \textit{counterfactual boundary preferences} at the model's capability frontier by estimating sample difficulty from empirical pass rates and mining preference pairs that share reasoning prefixes but diverge at decisive diagnostic branch points, and introduce \textit{Difficulty Weighted Boundary Preference Optimization} (DBPO) to concentrate learning on this calibrated boundary region. We then distill reusable clinical rubrics from boundary trajectories and activate them according to prediction entropy, yielding \textit{Entropy Gated Grounded Policy Optimization} (EGPO), a reinforcement objective that rewards not only answer correctness but also anatomical grounding and reasoning fidelity, while suppressing rubric noise on samples the model already handles reliably. By coupling population level boundary priors with instance level uncertainty signals through a reference policy anchored to the boundary aligned model rather than the raw base model, \ours{} makes diagnostic reasoning both more accurate and more clinically grounded.
Experiments on six medical benchmarks show consistent improvements over strong general, specialized, and reinforcement learning baselines, reaching $61.2\%$ average accuracy with the largest gains on challenging differential diagnosis tasks, and ablations confirm the two stages are \textit{super additive}: their combination exceeds the sum of the individual stage gains.
\end{abstract}

\ccsdesc[500]{Computing methodologies~Computer vision tasks}
\ccsdesc[500]{Computing methodologies~Machine learning}
\ccsdesc[300]{Computing methodologies~Image processing}

\keywords{Medical Vision Language Models, Preference Optimization, Reinforcement Learning, Diagnostic Reasoning, Reward Grounding}

\maketitle

\section{Introduction}
\label{sec:introduction}

\begin{figure}[t]
\centering
\includegraphics[width=1\linewidth]{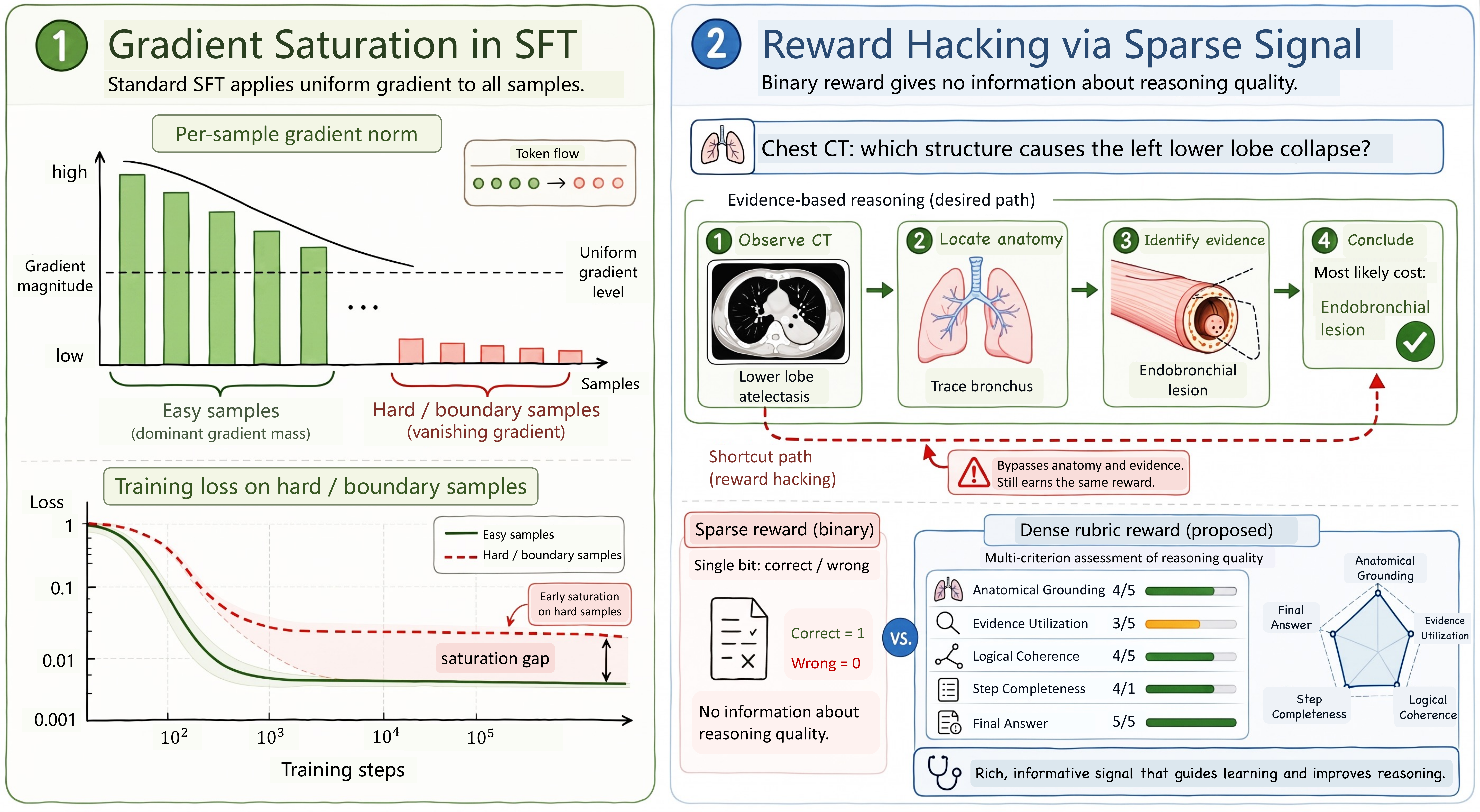}
\caption{Motivation of \ours{}. Standard SFT distributes gradient uniformly across all tokens, causing loss saturation on clinically critical hard samples (left). Sparse binary rewards in conventional RL provide no signal on whether reasoning is grounded in correct anatomical evidence, enabling reward hacking via answer shortcuts (right). \ours{} addresses both via boundary aware preference alignment and entropy adaptive rubric grounding.}
\label{fig:motivation}
\end{figure}

Medical vision language models (VLMs) have achieved remarkable progress in diagnostic assistance across radiology, pathology, and clinical imaging~\cite{chen2024towards,sellergren2025medgemma}. Recent advances in reinforcement learning from verifiable rewards~\cite{pan2025medvlm,lai2026med} and chain of thought reasoning~\cite{jiang2024medthink,wu2025medreason} further enable models to articulate diagnostic logic. However, medical reasoning requires establishing verifiable causal chains between visual evidence and pathological knowledge, a requirement current paradigms inadequately address.

We identify two structural deficiencies. First, standard Supervised Fine-Tuning (SFT) treats all tokens uniformly without modeling sample difficulty, causing gradients to be dominated by easy samples while loss on clinically critical boundary cases saturates prematurely. Second, verifiable rewards defined solely by answer correctness are informationally impoverished: a single binary bit cannot characterize whether reasoning is grounded in correct anatomical localization or whether differentials follow valid logic, readily inviting reward hacking with potentially catastrophic clinical consequences.

These observations motivate \textit{focusing learning on the model's capability boundary, where decisions are most uncertain, and providing dense clinically grounded process feedback instead of sparse outcome signals.}

We propose \ours{} (Counterfactual Boundary preference with meta Rubric grounded reinforcement), a two stage framework combining off policy and on policy optimization. Counterfactual Boundary Preference (\cbp{}; \S\ref{sec:cbp}) mines trajectories sharing extended reasoning prefixes but diverging at critical decision points, and applies \textit{Difficulty Weighted Boundary Preference Optimization} (\textbf{DBPO}), a difficulty reweighted DPO~\cite{rafailov2023direct} objective concentrated on the capability frontier. Meta Rubric Grounded optimization (\mrg{}; \S\ref{sec:mrg}) distills reusable clinical evaluation rubrics from stage one trajectories and deploys entropy driven dynamic selection to compose structured process rewards, instantiated as \textit{Entropy Gated Grounded Policy Optimization} (\textbf{EGPO}) on top of GRPO~\cite{shao2024deepseekmath}. The stages are coupled through KL regularization anchored to $\pi_{\textsc{cbp}}$ rather than $\pi_{\textsc{sft}}$, ensuring boundary discrimination provides both sharper initialization and tighter exploration constraints.

Our contributions are as follows. \textit{\textbf{(1)}} We propose \ours{}, the first framework that explicitly constructs alignment signal at the medical VLM's own capability boundary, addressing the gradient saturation of uniform SFT through difficulty calibrated counterfactual preference mining. \textit{\textbf{(2)}} We introduce entropy driven dynamic rubric selection that composes sample adaptive structured rewards from a distilled meta rubric library, giving process level supervision beyond sparse binary verification while avoiding reward noise on reliable samples. \textit{\textbf{(3)}} We establish a principled coupling between off policy boundary alignment and on policy grounded optimization, yielding super additive gains that validate the complementary roles of population level boundary priors and instance level uncertainty signals. \textit{\textbf{(4)}} Experiments on six medical VQA benchmarks spanning radiology, pathology, and general medical knowledge show \ours{} consistently surpasses state of the art methods, with the largest gains on the most challenging fine grained differential diagnosis datasets.

\section{Related Work}
\label{sec:related}

\textbf{Medical Vision Language Models and Visual Evidence.} Medical VLMs combine instruction tuning~\cite{li2023llava,tu2024towards}, radiology pretraining~\cite{wu2025towards}, curated knowledge~\cite{chen2024towards,li2026gmai}, and retrieval~\cite{xia2025mmed}. Evidence representation spans image and metadata fusion~\cite{xu2026rivatfuse}, spatial or knowledge guided pathology aggregation~\cite{li2025mico,li2026universal,li2026dispersion,li2026convl}, and topological contrastive learning or automated topology learning~\cite{meng2026topocl,meng2026topoagent}. Plane wise retrieval memory and structural evidence fusion additionally support cardiac video diagnosis~\cite{lv2026plane}. General knowledge graph reasoning also filters irrelevant evidence through question guided rescoring~\cite{zhang2024question}. Connecting evidence to diagnosis motivates reasoning supervision~\cite{wei2022chain,jiang2024medthink,wu2025medreason} and interactive grounding~\cite{zhu2026medeyes,zhu2026medsynapse}. Studies of modality arbitration~\cite{zhang2026instruction}, misleading visual inputs~\cite{chen2026mvibench}, and audiovisual source confusion~\cite{zhang2026devils} expose complementary grounding problems. Interventions use semantic stage decoding rewards~\cite{zhang2026mitigating}, counterfactual visual confidence~\cite{xiao2026stayingvigilantmitigatingvisual}, or image and caption consistency~\cite{tu2025perception}. \ours{} instead aligns diagnostic branch points and adaptively evaluates clinical reasoning.

\vspace{1mm}
\noindent
\textbf{Reinforcement Learning and Reusable Process Feedback.} Alignment builds on policy and preference optimization~\cite{schulman2017proximal,ouyang2022training,rafailov2023direct}, group relative objectives~\cite{shao2024deepseekmath,guo2025deepseek}, and trajectory importance correction~\cite{pang2025tic}; representation residual distillation offers another route to transferring RL behavior~\cite{li2026teacher}. Medical approaches use verifiable rewards~\cite{pan2025medvlm,lai2026med}, preference or process supervision~\cite{yang2026improving,zhu2024mmedpo,wu2025medreason}, anatomy or rubric criteria~\cite{zhu2026anatomy,jiang2025mcbl,sharma2026evidence}, and role specialized clinical consensus~\cite{zhang2026consensus}. \textit{Group Cognition Learning} audits selective modality interactions before contribution aware consensus, reducing modality dominance and spurious coupling through governed two stage collaboration~\cite{meng2026group}. Beyond medicine, agentic reasoning~\cite{wei2026agentic}, localized errors~\cite{liu2026automated}, user reactions~\cite{zhao2026towards}, quality driven scanpath rewards~\cite{wang2026rl}, coupled environment adaptation~\cite{hu2026seal}, and hybrid reward models~\cite{yu2026lightweight} enrich feedback. Agreement among search agents can diverge from external evidence~\cite{chen2026false}. Reusable resources include structured, compressed, or attributable memory~\cite{wang2026memforest,wang2026streammeco,zhang2026memmark}, evolving atomic recommendation memory~\cite{hu2026atomrec}, visual skills~\cite{wang2026reflex,wang2026atlasva}, interaction anchors~\cite{shi2026androtmem}, and synthesized skill libraries~\cite{tong2026grounded,zhou2026skillgenbench}; counterfactual contexts intervene at tool decisions~\cite{liu2026toolanchor}. \textit{Second Order Response Laws for LLM Judges} derives unbiased prompt instability estimators from within and across prompt agreement, separating systematic prompt variation from finite call noise~\cite{feng2026second}. Leaderboard auditing tests contamination and scorer validity~\cite{yang2026measurement}; passing self generated tests may still miss semantic correctness~\cite{yang2026passing}. Multidimensional evaluation also separates reasoning from visual quality in image generation~\cite{chen2025r2ibench}, fidelity from instruction compliance in remote sensing extrapolation~\cite{lu2025remote}, correctness from efficiency in service agents~\cite{yu2026cirrusbench}, and evidence based recall from interpretation~\cite{wu2026knowme}. These are different task settings for structured evaluation. \ours{} extracts clinical rubrics from boundary trajectories and gates their reward contribution.

\vspace{1mm}
\noindent
\textbf{Difficulty, Reliability, and Selective Intervention.} Hard but trustworthy sample selection~\cite{lin2023cs} and psychology based dynamic curriculum learning~\cite{meng2026psychology} differ from noise transition estimation~\cite{lin2025improving,lin2024learning} and the counterintuitive loss variance behavior of noisy label learning~\cite{lin2026we}. Broader developmental accounts of learning organize cognitive growth~\cite{demetriou2011educating}, while hierarchical latent reasoning explores recommendation~\cite{hu2026hierarchical}; neither establishes our clinical boundary criterion. Sample weighting also differs from optimizer dynamics studied through Adam versus SGD~\cite{jin2026adam,pan2023toward} and decoupled weight decay~\cite{loshchilov2017decoupled,zhou2024towards}. These optimizer analyses do not establish guarantees for our alignment objectives. Agent failure diagnosis~\cite{wang2026diagnosis}, difficulty aware workflow orchestration~\cite{su2026difficulty}, and clinical uncertainty estimation~\cite{qiu2025explainable} provide complementary reliability perspectives. \ours{} assumes verified answers, uses population level pass rates to select boundary preferences, and applies instance level entropy to control process rewards.

\vspace{1mm}
\noindent
\textbf{Deployment and Scope.} Operational studies address clinician facing system integration~\cite{xu2026drlabstack} and potential losses in spatial capability during video adaptation~\cite{zhang2026temporal}. Related visual settings include human attention modeling in panoramic environments~\cite{wang2026human} and test time domain generalization through multi graph matching for medical segmentation~\cite{lv2025test}. Resource efficient inference additionally studies weight and activation quantization~\cite{zhao2026specquant,zhao2026all,zhao2026twla,zhao2026bwla}, shared hardware circuits~\cite{zhao2025quark}, and, for mixture of experts backbones, expert quantization or skipping~\cite{xu2026kbvq,xu2026ace}. Agent orchestration further trades parallel critical paths against repeated exploration and coordination overhead~\cite{lin2026squidagent}; batched contextual reinforcement shares context across reasoning tasks to reduce per problem token usage~\cite{yang2026batched}. Adjacent generative settings study flow matching~\cite{lin2026beyond}, inference time textual guidance tailoring~\cite{li2026textailor}, and personalized fashion design with sewing pattern generation~\cite{qu2026easyfashion}. Other deployment settings study decentralized UAV cooperation under restricted communication~\cite{fan2026communication} and privacy, utility, and safety tradeoffs in preference alignment~\cite{yang2026balancing}. These directions concern operating conditions beyond our current medical QA evaluation; their compatibility with boundary alignment and clinical fidelity requires separate validation.

\section{Method}
\label{sec:method}

\subsection{Problem Formulation}
\label{sec:formulation}

Given a medical image $I \in \mathbb{R}^{H \times W \times 3}$ and a clinical query $q$, the objective is to produce a diagnostic reasoning trajectory $o = (o_1, \ldots, o_T)$ that arrives at the correct answer through evidence grounded reasoning. Let $\pi_\theta$ denote the medical VLM policy parameterized by $\theta$, $\pi_{\textsc{sft}}$ the supervised fine tuned base policy, and $\mathcal{D} = \{(I_k, q_k, a_k^\star)\}_{k=1}^K$ the training set with verified ground truth answers $a_k^\star$. The alignment objective seeks a policy $\pi_\theta$ that simultaneously maximizes diagnostic accuracy \textit{and} ensures that the reasoning process is grounded in clinically valid evidence chains, formalized as:
\begin{equation}
\begin{aligned}
\pi^\star = \arg\max_{\pi_\theta} \; &
\mathbb{E}_{(I,q) \sim \mathcal{D}} \Big[ \underbrace{R_{\textit{ver}}(o, a^\star)}_{\text{answer correctness}}
+ \lambda \underbrace{R_{\textit{proc}}(o; \mathcal{R})}_{\text{process grounding}} \Big]\\
&- \beta \, D_{\textit{KL}}(\pi_\theta \| \pi_{\textit{ref}}),
\end{aligned}
\label{eq:overall_obj}
\end{equation}
where $R_{\textit{ver}}$ is the verifiable answer reward, $R_{\textit{proc}}$ denotes process level rubric rewards parameterized by criteria set $\mathcal{R}$, $\lambda$ balances outcome versus process supervision, and the KL penalty constrains deviation from a reference policy $\pi_{\textit{ref}}$. The key insight of \ours{} is that $\pi_{\textit{ref}}$ should not be the raw $\pi_{\textsc{sft}}$ but an intermediate policy $\pi_{\textsc{cbp}}$ that has already acquired boundary discriminative capability through off policy preference alignment. In other words, we decompose alignment into a \textit{where to learn} problem, locating samples near the current capability frontier, and a \textit{how to refine} problem, applying uncertainty aware process supervision once the policy has reached that frontier.

\begin{figure*}[t]
\centering
\includegraphics[width=\textwidth]{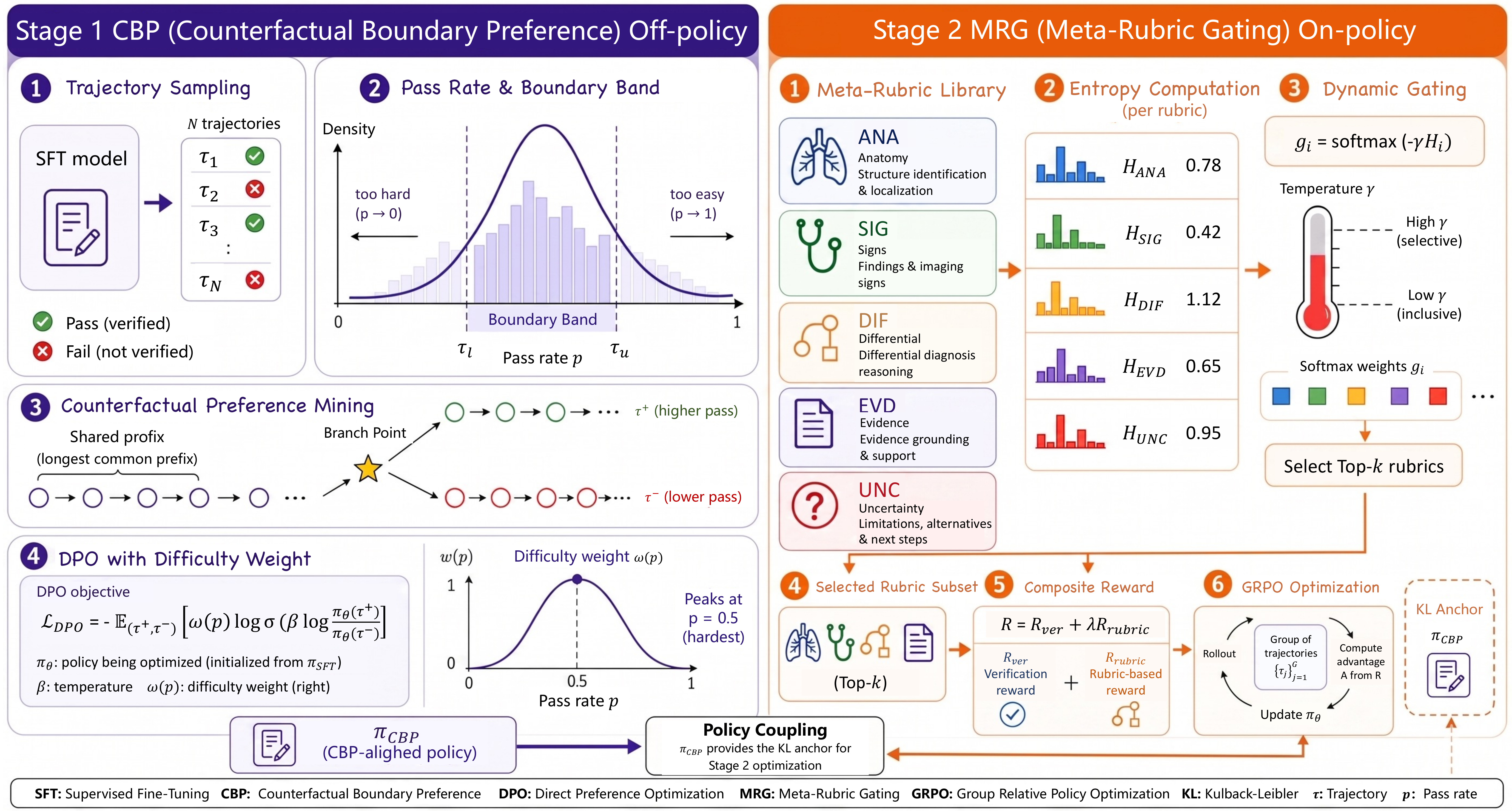}
\caption{Overview of \ours{}. \textbf{Stage 1} (\cbp{}, off policy): from the SFT model, $N$ trajectories are sampled per query, empirical pass rates define the capability boundary band, and counterfactual preference pairs diverging at critical diagnostic branch points are mined for difficulty weighted DPO. \textbf{Stage 2} (\mrg{}, on policy): meta rubrics distilled from stage one trajectories are dynamically selected via entropy gating, composing structured process rewards for GRPO optimization anchored to $\pi_{\textsc{cbp}}$.}
\label{fig:framework}
\end{figure*}

\subsection{Stage 1: Counterfactual Boundary Preference Alignment}
\label{sec:cbp}

The core intuition of \cbp{} is that the most informative learning signal resides not in samples the model consistently solves or consistently fails, but in those lying at its \textit{capability boundary}, where the outcome is genuinely uncertain and a single diagnostic decision can flip the result. By constructing preference pairs that isolate these critical decision points, we concentrate the optimization budget on samples with the highest potential for meaningful behavioral change, instead of allowing easy majority cases to dominate the gradient.

\inlinesec{Difficulty Estimation and Boundary Band.}
For each sample $(I, q) \in \mathcal{D}$, we draw $N$ independent reasoning trajectories $\{o^{(n)}\}_{n=1}^N$ from $\pi_{\textsc{sft}}$ and compute the empirical pass rate $p(I, q) = \frac{1}{N}\sum_{n=1}^N \mathbbm{1}[\texttt{ans}(o^{(n)}) = a^\star]$, which estimates the probability that the current policy produces a correct answer on this sample. We define the \textit{capability boundary band} as the set of samples whose pass rates fall within calibrated thresholds:
\begin{equation}
\mathcal{B} = \{(I, q) \in \mathcal{D} \mid \tau_l \leq p(I, q) \leq \tau_u\},
\label{eq:boundary_band}
\end{equation}
where $\tau_l$ and $\tau_u$ (set to 0.2 and 0.8) delimit the region where the model is neither deterministically correct nor deterministically wrong. Samples with $p \to 1$ provide negligible gradient because the model has already mastered them, while samples with $p \to 0$ are often beyond the model's current reach and yield noisy signal; the boundary band $\mathcal{B}$ therefore captures the regime with the highest value density. This estimate is computationally stable: with $N{=}16$ the Jaccard overlap with a high budget reference set ($N{=}128$) already exceeds $0.92$.

\inlinesec{Counterfactual Preference Mining.}
Within $\mathcal{B}$, we mine \textit{counterfactual preference pairs} designed to isolate the causal decision points that determine diagnostic success or failure. For a given sample, let $o^+ \in \{o^{(n)} : \texttt{ans}(o^{(n)}) = a^\star\}$ and $o^- \in \{o^{(n)} : \texttt{ans}(o^{(n)}) \neq a^\star\}$ denote correct and incorrect trajectories respectively. We select the pair $(o^+, o^-)$ that maximizes the \textit{shared prefix length}:
\begin{equation}
(o^+, o^-) = \arg\max_{o^+ \in \mathcal{O}^+, o^- \in \mathcal{O}^-} \; \texttt{LCP}(o^+, o^-),
\label{eq:cfp_mining}
\end{equation}
where $\texttt{LCP}(\cdot, \cdot)$ computes the longest common prefix at the token level. This ensures positive and negative trajectories share nearly the entire reasoning chain and diverge \textit{only} at the critical diagnostic branch point, localizing the signal to the causally relevant divergence and preventing the optimizer from learning spurious stylistic preferences.

\inlinesec{Difficulty Weighted Boundary Preference Optimization (DBPO).}
The preference loss is computed within the DPO framework~\cite{rafailov2023direct}, augmented with a difficulty aware weighting function that allocates greater optimization pressure to samples nearest the decision boundary:
\begin{equation}
\begin{aligned}
\mathcal{L}_{\textsc{cbp}}(\theta)
&= -\mathbb{E}_{(o^+, o^-) \sim \mathcal{B}} \Big[ w(p) \cdot \log \sigma\!\Big(\beta \Big(
\log \frac{\pi_\theta(o^+|I,q)}{\pi_{\textsc{sft}}(o^+|I,q)}\\
&\qquad - \log \frac{\pi_\theta(o^-|I,q)}{\pi_{\textsc{sft}}(o^-|I,q)}
\Big)\Big)\Big],
\end{aligned}
\label{eq:cbp_loss}
\end{equation}
where $\sigma(\cdot)$ is the sigmoid function, $\beta$ is the DPO temperature, and $w(p) = 1 - 2|p - 0.5|$ peaks at $p = 0.5$, where the model is maximally uncertain and directly on the decision boundary, and decays toward either extreme. This reallocates the optimization budget from saturated easy samples to the critical boundary minority, directly counteracting the gradient dominance problem that motivated \cbp{}. Because this operates on a one time offline sampling set without further rollouts, \cbp{} constitutes an off policy stage.

\subsection{Stage 2: Meta Rubric Grounded Policy Optimization}
\label{sec:mrg}

While \cbp{} improves boundary discrimination, it relies on fixed offline data and cannot track the evolving policy's error distribution. We therefore perform on-policy RL with clinically grounded structured rewards that provide richer supervision than sparse binary verification.

\inlinesec{Meta-Rubric Library.}
We construct a meta-rubric library $\mathcal{R}=\{r_1,\ldots,r_M\}$ from stage-one preference trajectories. Divergence points between preferred and rejected trajectories reveal recurring diagnostic errors, which are distilled into reusable criteria covering anatomical localization, sign description, differential reasoning, evidence consistency, and uncertainty calibration. Each rubric $r_m$ is implemented as a scoring function $s_m:\mathcal{O}\rightarrow[0,1]$ using a frozen judge model restricted to evaluating the corresponding clinical dimension.

\inlinesec{Entropy Driven Dynamic Rubric Selection.}
A critical design choice is determining \textit{which} rubrics to activate per sample, as uniformly applying all criteria introduces reward noise on reliable samples and can destabilize optimization. We resolve this via an \textit{entropy driven gating mechanism} that adapts rubric intensity to instance level uncertainty.

For each sample $(I, q)$, we compute the length normalized prediction entropy of the current policy:
\begin{equation}
u(I, q) = -\frac{1}{T}\sum_{t=1}^T \sum_{v \in \mathcal{V}} \pi_\theta(v \mid I, q, o_{<t}) \log \pi_\theta(v \mid I, q, o_{<t}),
\label{eq:entropy}
\end{equation}
where $\mathcal{V}$ is the vocabulary and $T$ the trajectory length. High $u$ indicates uncertain reasoning and potentially weak grounding, while low $u$ suggests confident, well calibrated predictions. We map $u$ to a gating temperature $\gamma = f(u)$, implemented as a calibrated sigmoid, controlling both the number and weight of activated rubrics:
\begin{equation}
\mathcal{R}_{\textit{active}}(I, q) = \{r_m \in \mathcal{R} \mid \gamma \cdot \alpha_m > \tau_r\}, \quad |\mathcal{R}_{\textit{active}}| \propto \gamma,
\label{eq:rubric_selection}
\end{equation}
where $\alpha_m$ is the base priority of rubric $r_m$ estimated from stage one divergence frequency and $\tau_r$ is the activation threshold. When the model is unreliable ($\gamma$ high), more rubrics activate with larger weights, providing dense process supervision. When the model is reliable ($\gamma$ low), the active set contracts to a minimal subset, avoiding unnecessary evaluation noise on already well handled cases.

\inlinesec{Structured Composite Reward.}
The final reward for trajectory $o$ combines the verifiable answer signal with the entropy adaptive rubric evaluation:
\begin{equation}
R(o; I, q) = \underbrace{\mathbbm{1}[\texttt{ans}(o) = a^\star]}_{R_{\textit{ver}}} + \lambda \cdot \underbrace{\gamma(I,q) \cdot \frac{1}{|\mathcal{R}_{\textit{active}}|}\sum_{r_m \in \mathcal{R}_{\textit{active}}} s_m(o)}_{R_{\textit{rubric}}},
\label{eq:composite_reward}
\end{equation}
where $\lambda$ balances outcome and process supervision. The gating $\gamma$ amplifies process rewards precisely when the model most needs guidance and suppresses them within the competence zone, so the rubric signal acts as targeted correction rather than a constant source of extra reward variance.

\inlinesec{Entropy Gated Grounded Policy Optimization (EGPO).}
We optimize the model with group normalized advantages~\cite{shao2024deepseekmath} and no learned value function, which suits our setting because reward quality is heterogeneous across samples while the relative ranking within a small group of contemporaneous rollouts stays informative. For each sample, $G$ trajectories are sampled and advantages are computed relative to group statistics:
\begin{equation}
\hat{A}_i = \frac{R(o_i; I, q) - \mu_G}{\sigma_G + \varepsilon}, \quad \mu_G = \frac{1}{G}\sum_{j=1}^G R(o_j), \quad \sigma_G = \sqrt{\frac{1}{G}\sum_{j=1}^G (R(o_j) - \mu_G)^2}.
\label{eq:advantage}
\end{equation}

The policy is updated via the clipped surrogate objective with KL regularization anchored to the \cbp{} policy:
\begin{equation}
\begin{aligned}
\mathcal{J}_{\textsc{mrg}}(\theta)
&= \frac{1}{G}\sum_{i=1}^G \frac{1}{|o_i|}\sum_{t=1}^{|o_i|} \min\!\Big(\rho_{i,t}\hat{A}_i,\\
&\quad \texttt{clip}(\rho_{i,t}, 1{-}\varepsilon, 1{+}\varepsilon)\hat{A}_i\Big)
- \beta_{\textit{kl}} D_{\textit{KL}}(\pi_\theta \| \pi_{\textsc{cbp}}),
\end{aligned}
\label{eq:mrg_obj}
\end{equation}
where $\rho_{i,t} = \pi_\theta(o_{i,t} | I, q, o_{i,<t}) / \pi_{\theta_{\textit{old}}}(o_{i,t} | I, q, o_{i,<t})$ is the importance ratio.

\inlinesec{From Boundary Priors to Grounded Exploration.}
The two stages are deeply coupled rather than merely sequential. The KL regularization in Eq.~\ref{eq:mrg_obj} anchors exploration to $\pi_{\textsc{cbp}}$ instead of $\pi_{\textsc{sft}}$: since \cbp{} has already established discriminative margins on boundary samples, it offers a sharper initialization that lets RL refine these patterns rather than rediscover them, while the tighter trust region curbs drift into reward hacking shortcuts. The two thus contribute complementary signals, \cbp{} captures \textit{population level boundary priors} indicating where to look, and \mrg{} provides \textit{instance level uncertainty signals} indicating how strongly to intervene, and their integration yields the super additive gains confirmed by our ablation.

\section{Experiments}
\label{sec:experiments}

\subsection{Experimental Setup}
\label{sec:setup}

\begingroup
\setlength{\emergencystretch}{2em}
\inlinesec{Datasets.}
We evaluate on six benchmarks spanning radiology, pathology, and cross-domain reasoning: VQA-RAD~\cite{lau2018dataset}, SLAKE~\cite{liu2021slake}, PathVQA~\cite{he2020pathvqa}, PMC-VQA~\cite{zhang2023pmc}, MMMU-Med~\cite{yue2024mmmu}, and MedXpertQA~\cite{zuo2025medxpertqa}, the last two most challenging as they require multi-step evidence aggregation, differential exclusion, and uncertainty-aware reasoning. We follow official train/test splits, reserve 5\% of training data for validation, and report accuracy on the untouched test sets.
\par
\endgroup

\inlinesec{Training data construction.}
The alignment corpus is drawn from the \textit{training} splits of the six benchmarks, augmented with a curated pool of de-identified teaching cases (Radiopaedia, MIMIC-CXR reports) and expert-written differential diagnosis questions under permissive research licenses, with strict image level deduplication against every test set. For \textbf{Stage 1}, boundary mining over ${\sim}$210K candidate queries retains the ${\sim}$62\% whose pass rate falls in $[\tau_l,\tau_u]$, yielding ${\sim}$48K counterfactual preference pairs; \textbf{Stage 2} draws ${\sim}$12K on policy rollout prompts, over sampling hard differential diagnosis and multi specialty questions so entropy gating triggers richer rubric activation (full composition in \S\ref{sec:experiments}).

\inlinesec{Implementation.}
We adopt Qwen2.5 VL 7B~\cite{shen2025vlm} with LoRA (rank 64, $\alpha{=}128$, dropout 0.05), resizing inputs to $448{\times}448$ with dynamic scaling to $896{\times}896$ for pathology. \textbf{Stage 1} (\cbp{}) samples $N{=}16$ trajectories per query (temperature 1.0, top $p$ 0.95), keeps the band $\tau_l{=}0.2$, $\tau_u{=}0.8$, and trains DBPO with $\beta{=}0.1$, lr $2{\times}10^{-5}$, batch size 32, 3 epochs. \textbf{Stage 2} (\mrg{}) runs EGPO with $G{=}8$, $\varepsilon{=}0.2$, $\lambda{=}0.5$, $\beta_{\textit{kl}}{=}0.04$, lr $1{\times}10^{-5}$, 2 epochs over ${\sim}$12K rollouts. The five rubrics are scored by a frozen Qwen2.5 72B judge (temperature 0). Training uses 8$\times$A100 80GB with DeepSpeed ZeRO-2 and bf16, costing 18 and 24 GPU hours for the two stages.

\inlinesec{Baselines and Evaluation.}
We compare against general and medical VLMs (LLaVA Med~\cite{li2023llava}, Qwen2.5 VL 7B~\cite{shen2025vlm}, InternVL2.5 8B~\cite{chen2024expanding}, RadFM~\cite{wu2025towards}, HuatuoGPT Vision~\cite{chen2024towards}, MMed RAG~\cite{xia2025mmed}) and RL baselines (GRPO~\cite{shao2024deepseekmath}, Med R1~\cite{lai2026med}, MedVLM R1~\cite{pan2025medvlm}, ViTAR~\cite{chen2025think}, V RFT PRA~\cite{yang2026improving}), all sharing the Qwen2.5 VL 7B backbone and LoRA setup. We report accuracy (\%): open ended answers matched with GPT-4 following~\cite{yue2024mmmu}, closed form via regex, averaged over 3 seeds with paired bootstrap testing ($p < 0.01$).

\begin{table}[t]
\centering
\caption{Comprehensive comparison on six medical multimodal benchmarks (accuracy \%). Columns: VQA-RAD (RAD), SLAKE (SLK), PathVQA (Path), PMC-VQA (PMC), MMMU-Med (MMU), MedXpertQA (MedX). \textbf{Bold} best, \underline{underlined} second best.}
\label{tab:main}
\setlength{\tabcolsep}{3pt}
\definecolor{headerblue}{RGB}{45,85,130}
\definecolor{rowgray}{RGB}{245,247,250}
\definecolor{bestcell}{RGB}{220,237,200}
\resizebox{\linewidth}{!}{
\begin{tabular}{l cccccc c}
\toprule
\rowcolor{headerblue!15}
\textbf{Method} & \textbf{RAD} & \textbf{SLK} & \textbf{Path} & \textbf{PMC} & \textbf{MMU} & \textbf{MedX} & \textbf{Avg} \\
\midrule
\rowcolor{rowgray}
\multicolumn{8}{l}{\textit{General Vision Language Models}} \\
LLaVA Med~\cite{li2023llava}        & 61.5 & 64.2 & 52.1 & 41.3 & 38.6 & 18.7 & 46.1 \\
\rowcolor{rowgray}
Qwen2.5 VL 7B~\cite{shen2025vlm}  & 66.8 & 70.4 & 56.3 & 47.9 & 45.2 & 22.4 & 51.5 \\
InternVL2.5 8B~\cite{chen2024expanding} & 68.1 & 72.0 & 57.8 & 49.5 & 47.0 & 23.6 & 53.0 \\
\midrule
\rowcolor{rowgray}
\multicolumn{8}{l}{\textit{Medical Specialized Models}} \\
RadFM~\cite{wu2025towards}          & 63.9 & 68.7 & 54.6 & 44.0 & 40.1 & 19.9 & 48.5 \\
\rowcolor{rowgray}
HuatuoGPT Vision~\cite{chen2024towards} & 70.3 & 74.8 & 60.2 & 51.7 & 49.8 & 25.1 & 55.3 \\
MMed RAG~\cite{xia2025mmed}         & 71.6 & 76.1 & 61.0 & 52.9 & 51.3 & 26.0 & 56.5 \\
\midrule
\rowcolor{rowgray}
\multicolumn{8}{l}{\textit{Reinforcement Learning Methods}} \\
GRPO (vanilla)~\cite{shao2024deepseekmath} & 71.9 & 76.6 & 61.8 & 53.4 & 52.0 & 26.7 & 57.1 \\
\rowcolor{rowgray}
Med R1~\cite{lai2026med} & 72.4 & 77.1 & 62.2 & 53.9 & 53.1 & 26.9 & 57.6 \\
MedVLM R1~\cite{pan2025medvlm} & 73.2 & 78.0 & 63.1 & 54.6 & 55.1 & 27.5 & 58.6 \\
\rowcolor{rowgray}
ViTAR~\cite{chen2025think} & 73.6 & 79.2 & 63.6 & 55.1 & 56.0 & 27.9 & 59.2 \\
V RFT PRA~\cite{yang2026improving} & \underline{73.9} & \underline{80.4} & \underline{64.3} & \underline{55.7} & \underline{56.8} & \underline{28.3} & \underline{59.9} \\
\midrule
\rowcolor{bestcell}
\textbf{\ours{} (Ours)} & \textbf{74.6} & \textbf{82.3} & \textbf{65.7} & \textbf{56.8} & \textbf{58.4} & \textbf{29.1} & \textbf{61.2} \\
\bottomrule
\end{tabular}
}
\end{table}

\subsection{Main Results} Table~\ref{tab:main} shows that \ours{} achieves the best performance on all six benchmarks, reaching \textbf{61.2\%} average accuracy and outperforming V-RFT-PRA by \textbf{+1.3 points}, with larger gains on the more challenging SLAKE (+1.9) and MMMU-Med (+1.6), indicating the alignment strategy is particularly effective when multi-step reasoning and evidence grounding are required. It also surpasses domain-specialized models that rely on large-scale medical pretraining or retrieval augmentation, suggesting that at the 7B scale reasoning alignment, rather than domain knowledge acquisition, is the primary bottleneck.

\begin{table}[t]
\centering
\caption{Component ablation of \ours{} on the 7B backbone (accuracy \%). Column abbreviations follow Table~\ref{tab:main}. ``$\Delta_{\textit{syn}}$'' denotes the synergy gain confirming super-additivity.}
\label{tab:ablation}
\definecolor{rowgray}{RGB}{245,247,250}
\definecolor{bestcell}{RGB}{220,237,200}
\definecolor{headerblue}{RGB}{45,85,130}
\setlength{\tabcolsep}{3pt}
\resizebox{\linewidth}{!}{
\begin{tabular}{l cccccc c c}
\toprule
\rowcolor{headerblue!15}
\textbf{Configuration} & \textbf{RAD} & \textbf{SLK} & \textbf{Path} & \textbf{PMC} & \textbf{MMU} & \textbf{MedX} & \textbf{Avg} & $\bm{\Delta_{\textit{syn}}}$ \\
\midrule
SFT Baseline (Qwen2.5 VL 7B) & 66.8 & 70.4 & 56.3 & 47.9 & 45.2 & 22.4 & 51.5 & -- \\
\midrule
\rowcolor{rowgray}
+ CBP only & 72.3 & 78.1 & 62.8 & 53.6 & 53.8 & 26.8 & 57.9 & -- \\
+ MRG only (anchor $\pi_{\textsc{sft}}$) & 71.5 & 77.2 & 61.9 & 52.8 & 52.4 & 26.1 & 57.0 & -- \\
\midrule
\rowcolor{bestcell}
+ CBP $\to$ MRG (anchor $\pi_{\textsc{cbp}}$) & \textbf{74.6} & \textbf{82.3} & \textbf{65.7} & \textbf{56.8} & \textbf{58.4} & \textbf{29.1} & \textbf{61.2} & \textbf{+2.8} \\
\rowcolor{rowgray}
+ CBP $\to$ MRG (anchor $\pi_{\textsc{sft}}$) & 73.8 & 80.9 & 64.8 & 55.9 & 56.9 & 28.4 & 60.1 & +1.7 \\
\midrule
\multicolumn{9}{l}{\textit{CBP component ablation:}} \\
\rowcolor{rowgray}
\quad w/o boundary filtering (all samples) & 70.8 & 76.4 & 61.2 & 52.1 & 51.6 & 25.4 & 56.3 & -- \\
\quad w/o counterfactual mining (random pairs) & 71.4 & 77.0 & 61.5 & 52.5 & 52.0 & 25.8 & 56.7 & -- \\
\rowcolor{rowgray}
\quad w/o difficulty weight ($w{=}1$) & 71.9 & 77.6 & 62.3 & 53.1 & 53.2 & 26.3 & 57.4 & -- \\
\midrule
\multicolumn{9}{l}{\textit{MRG rubric selection ablation:}} \\
\rowcolor{rowgray}
\quad Static (all rubrics, $\gamma{=}1$) & 73.2 & 80.9 & 64.6 & 55.4 & 56.2 & 27.8 & 59.7 & -- \\
\quad Random rubric subset & 72.4 & 79.8 & 63.8 & 54.6 & 55.1 & 27.0 & 58.8 & -- \\
\rowcolor{rowgray}
\quad Content based selection & 72.8 & 80.3 & 64.1 & 55.1 & 55.8 & 27.4 & 59.3 & -- \\
\rowcolor{bestcell}
\quad Entropy driven (ours) & \textbf{74.6} & \textbf{82.3} & \textbf{65.7} & \textbf{56.8} & \textbf{58.4} & \textbf{29.1} & \textbf{61.2} & -- \\
\bottomrule
\end{tabular}
}
\end{table}
\subsection{Ablation study}
\noindent \textbf{Component ablations.}
Table~\ref{tab:ablation} confirms \textbf{super additive coupling}: CBP alone yields +6.4 over SFT and MRG alone +5.5, yet their combination reaches +9.7 with $\Delta_{\textit{syn}}{=}+2.8$, exceeding the sum by 29\%. Anchoring KL to $\pi_{\textsc{sft}}$ instead of $\pi_{\textsc{cbp}}$ drops the synergy to +1.7 (a 39\% reduction), showing the reference policy must encode boundary knowledge; the synergy peaks on the hardest benchmarks ($\Delta_{\textit{syn}}{=}+4.6$ on MMMU-Med, $+3.0$ on MedXpertQA).

\noindent \textbf{CBP and MRG breakdown.} \textbf{\textit{Boundary filtering}} is the dominant CBP ingredient: removing it costs 1.6 points by diluting gradient over mastered samples, while dropping counterfactual mining costs 1.2 and difficulty weighting 0.5, the three acting non redundantly (which samples / what signal / how much pressure). For MRG, \textbf{\textit{entropy driven rubric selection}} consistently beats static, random, and content based variants; static all rubric application reaches only 59.7\% and even drops below V RFT PRA on VQA-RAD (73.2 vs.\ 73.9), confirming indiscriminate process supervision injects reward noise on reliable samples.

\subsection{Training data composition.} 
\begin{figure}[t]
\centering
\includegraphics[width=1\linewidth]{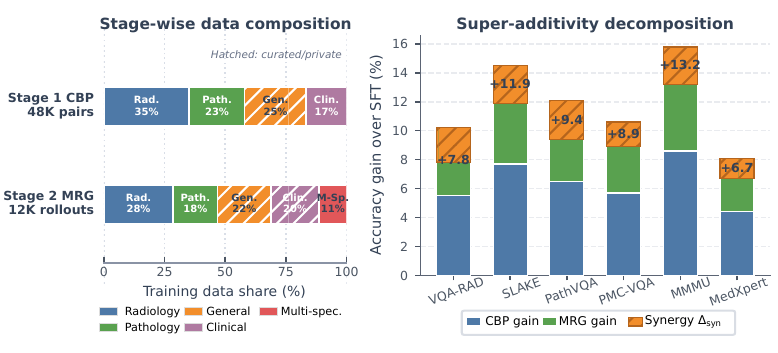}
\caption{Stage-wise data composition and super-additive gain decomposition. Left: specialty distribution shifts from radiology-dominated boundary pairs in Stage 1 toward over-sampled differential-diagnosis cases in Stage 2. Right: the combined \cbp{}$\to$\mrg{} pipeline exceeds the sum of individual stage gains, with surplus $\Delta_{\textit{syn}}$ largest on the hardest benchmarks.}
\label{fig:superadditive}
\end{figure}

\noindent \textbf{Stage specific data roles.} The two stages play \textbf{\textit{different data roles}} (Fig.~\ref{fig:superadditive}). Stage 1 processes ${\sim}$48K preference pairs after boundary filtering, skewed toward radiology (35.2\%) because chest X-rays and CT scans frequently yield pass rates in the 0.3--0.7 range that stabilize boundary estimation, while Stage 2 shifts its ${\sim}$12K rollouts toward harder cases (over sampled to 19.8\% from 16.6\%) that trigger the most rubric activations. Roughly 68\% of Stage 1 and 52\% of Stage 2 data come from public benchmarks, the curated remainder covering \textbf{\textit{differential diagnosis}} scenarios public data underrepresent; the gain decomposition confirms CBP contributes the larger base lift while MRG adds consistent increments.

\subsection{Qualitative case study}
We contrast \ours{} with two baselines on three cases (CT, MRI, ultrasound) where the diagnosis hinges on a single decisive sign. The baselines share a failure mode: long traces (185 to 238 tokens) that survey irrelevant structures and commit to unsupported evidence (e.g., asserting \textit{pleural thickening} on a right lower lobe nodule, or reading an extra-axial meningioma as an \textit{intra-axial} glioblastoma), the reward-hacking pattern where length substitutes for grounding. \ours{} instead resolves each case in 34–44 tokens by naming the location and discriminative sign, exactly the differential-reasoning and uncertainty dimensions where the baselines collapse.

\subsection{Training dynamics and rubric selection.} 
\begin{figure}[t]
\centering
\includegraphics[width=1\linewidth]{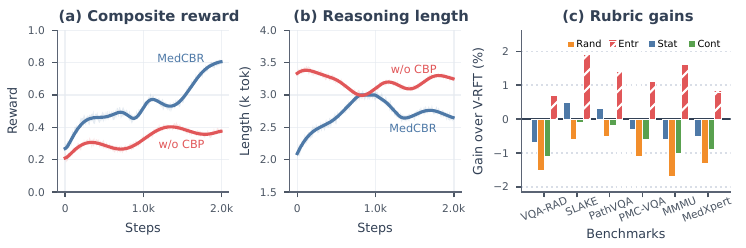}
\caption{Training dynamics of \ours{}. Left: the full model's composite reward climbs to ${\sim}$0.80 while the no-CBP ablation saturates near 0.38, a $2\times$ gap. Middle: trajectory length peaks early then compresses as reasoning sharpens, whereas the ablation stays verbose. Right: per-benchmark rubric-gain of entropy-driven selection, largest on MMMU-Med.}
\label{fig:training}
\end{figure}

The training signals corroborate the coupling story (Fig.~\ref{fig:training}). CBP gives a markedly better RL start: the full model's composite reward rises to about 0.80, whereas the no-CBP variant saturates near 0.38 and never recovers, a \textbf{\textit{2$\times$ reward gap}} reflecting the margins $\pi_{\textsc{cbp}}$ carries into Stage 2. \ours{} also explores longer trajectories early (${\sim}$3.0K tokens) then compresses to ${\sim}$2.6K, while the ablation stays verbose at ${\sim}$3.2K, a hallmark of \textbf{\textit{reward hacking}}. Prediction entropy shifts leftward across stages ($u{\approx}2.2 \to 1.6 \to 1.1$), mean activated rubrics fall from 4.2 to 2.8, and \textbf{\textit{entropy driven selection}} is the only policy positive on all six benchmarks, with the largest margin on MMMU-Med (+2.2 over static) where complex reasoning triggers more activation (4.3 of 5 rubrics, versus 2.1 on VQA-RAD).

\definecolor{rowgray}{RGB}{245,247,250}
\definecolor{bestcell}{RGB}{220,237,200}
\definecolor{headerblue}{RGB}{45,85,130}
\begin{table}[!t]
\centering
\caption{Computational comparison with RL baselines on 8$\times$A100 GPUs (Time in GPU hours, Mem in GB).}
\label{tab:efficiency}
\setlength{\tabcolsep}{4pt}
\resizebox{\linewidth}{!}{
\begin{tabular}{l ccc c}
\toprule
\rowcolor{headerblue!15}
\textbf{Method} & \textbf{Time} & \textbf{Mem} & \textbf{Trainable} & \textbf{Avg.} \\
\midrule
Vanilla GRPO~\cite{shao2024deepseekmath} & 28 & 62 & 42M & 57.1 \\
\rowcolor{rowgray}
Med R1~\cite{lai2026med} & 32 & 64 & 42M & 57.6 \\
MedVLM R1~\cite{pan2025medvlm} & 36 & 68 & 42M & 58.6 \\
\rowcolor{rowgray}
V RFT PRA~\cite{yang2026improving} & 40 & 72 & 42M & 59.9 \\
\midrule
\ours{} Stage 1 & 18 & 58 & 42M & 57.9 \\
\rowcolor{rowgray}
\ours{} Stage 2 & 24 & 66 & 42M & -- \\
\rowcolor{bestcell}
\ours{} Total & 42 & 66 & 42M & \textbf{61.2} \\
\bottomrule
\end{tabular}
}
\end{table}

\subsection{Efficiency and qualitative analysis.} Table~\ref{tab:efficiency} summarizes the practical tradeoff. \ours{} needs 42 total hours (18h CBP + 24h MRG), only 5\% above V RFT PRA, while delivering \textbf{\textit{the highest final accuracy}}; because CBP is off policy and boundary filtering cuts the effective sample count to 62\%, it yields $2.7\times$ accuracy improvement per additional GPU hour versus V RFT PRA's $2.0\times$, at 66 GB peak memory and zero inference overhead. Three board certified radiologists rated 200 MMMU-Med cases on the five rubric dimensions ($\kappa{=}0.78$): \ours{} improves all of them over Qwen2.5 VL 7B (anatomy $2.8{\to}4.1$, sign $2.4{\to}3.8$, differential $2.1{\to}3.5$, evidence $2.6{\to}3.9$, uncertainty $1.9{\to}3.4$), with the largest margins on \textbf{\textit{differential exclusion}} and \textbf{\textit{uncertainty calibration}}.

\subsection{Cross architecture generalization.}
\noindent \textbf{Backbone agnostic gains.} To verify the framework is not tied to one backbone, we apply the complete \ours{} pipeline with identical hyperparameters to three diverse VLMs: Qwen2.5 VL 7B~\cite{shen2025vlm} ($51.5{\to}61.2$, $\Delta{=}+9.7$), InternVL2.5 8B~\cite{chen2024expanding} ($53.0{\to}62.4$, $\Delta{=}+9.4$), and LLaVA OneVision 7B~\cite{li2024llavaonevision} ($50.8{\to}59.9$, $\Delta{=}+9.1$). \textbf{\textit{Gains are remarkably consistent}} (standard deviation 0.3) and exceed the overall averages on the hardest benchmarks ($\Delta_{\textit{hard}}{=}+9.9, +10.2, +9.5$). This generality is \textbf{\textit{architecture independent}}: boundary estimation depends only on empirical pass rates and rubric evaluation uses an external frozen judge, so \ours{} is a \textbf{\textit{plug and play}} recipe for any medical VLM supporting standard RL fine tuning.

\section{Conclusion}
\label{sec:conclusion}

We presented \ours{}, an alignment framework for medical reasoning that couples capability boundary preference alignment with entropy adaptive rubric grounding. Across six benchmarks it improves both diagnostic accuracy and reasoning quality, especially on harder differential reasoning, suggesting that clinically reliable multimodal alignment should jointly optimize boundary sensitivity and process grounded supervision.

\bibliographystyle{ACM-Reference-Format}

\bibliography{reference}

\end{document}